\documentclass[runningheads]{llncs}

\usepackage{eccv}

\usepackage{eccvabbrv}

\usepackage{graphicx}
\usepackage{booktabs}
\usepackage{tabularx}
\usepackage[table]{xcolor}
\usepackage{wrapfig}

\usepackage[accsupp]{axessibility}  % Improves PDF readability for those with disabilities.

\usepackage{hyperref}

\usepackage{orcidlink}

\begin{document}

\title{Humanoid Whole-Body Manipulation \\via Active Spatial Brain and \\Generalizable Action Cerebellum} 

\titlerunning{Humanoid Whole-Body Manipulation}

% Second Author\inst{2,3}\orcidlink{1111-2222-3333-4444} \and
% Third Author\inst{3}\orcidlink{2222--3333-4444-5555}}
\renewcommand{\thefootnote}{\fnsymbol{footnote}}
\author{
    Zhizhao Liang\inst{1}$^{*}$ \and
    Yi-Lin Wei\inst{1}\protect$^{*}$ \and
    Xuhang Chen\inst{1}\protect$^{*}$ \and
    Mu Lin\inst{1} \and
    Yi-Xiang He\inst{1} \and
    Zhexi Luo\inst{1} \and
    Jun-Hui Liu\inst{1,2} \and
    Kun-Yu Lin\inst{1} \and
    Wei-Shi Zheng\inst{1,2,3}$^{\dagger}$
}

\authorrunning{Z. Liang et al.}

\institute{
  School of Computer Science and Engineering, Sun Yat-sen University, China \and
  Peng Cheng Laboratory, China \and
  Key Laboratory of Machine Intelligence and Advanced Computing, Ministry of Education, China\\
  % {\small\textcolor{deepPink}{\href{https://leungchaos.github.io/Humanoid-Whole-Body-Manipulation-via-Active-Spatial-Brain-and-Generalizable-Action-Cerebellum/}{https://leungchaos.github.io/Humanoid-Whole-Body-Manipulation-via-Active-Spatial-Brain-and-Generalizable-Action-Cerebellum/}}}
  \url{https://leungchaos.github.io/Humanoid-Whole-Body-Manipulation-via-Active-Spatial-Brain-and-Generalizable-Action-Cerebellum/}\\
  \email{liangzhzh26@mail2.sysu.edu.cn}
}

% Springer Heidelberg, Tiergartenstr.~17, 69121 Heidelberg, Germany

\maketitle
\footnotetext[1]{Equal contribution.} 
\footnotetext[4]{Corresponding author.}

\begin{center}
  \includegraphics[width=\textwidth]{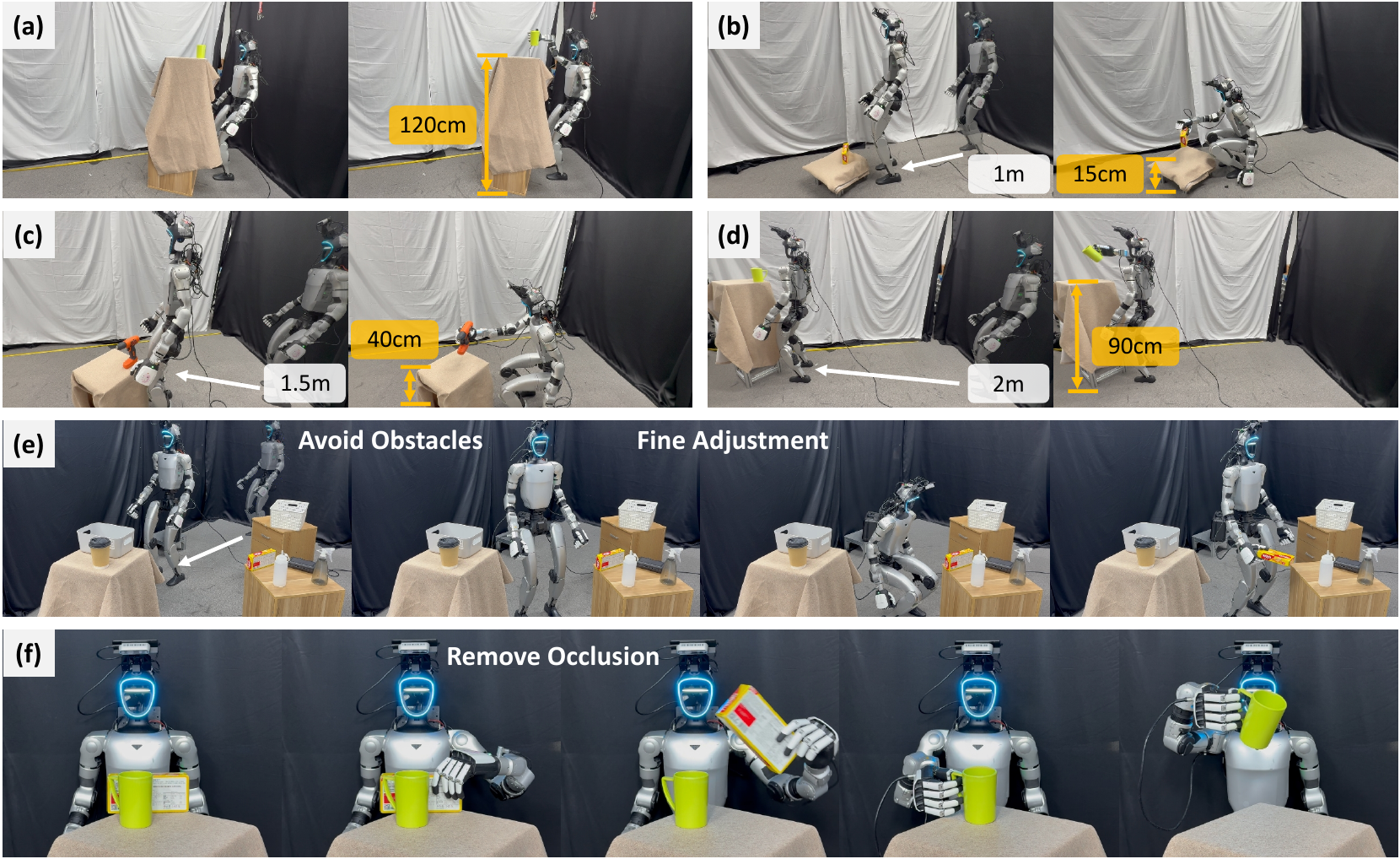}
  \captionof{figure}{This work enables generalizable humanoid whole-body manipulation in complex spatial environments through a multi-agent multimodal framework composed of an Active Spatial Brain and a Generalizable Action Cerebellum, without relying on task-specific data.}
  \label{fig:setting}
\end{center}

\begin{abstract}
In this paper, we explore spatial-aware humanoid whole-body manipulation task.
Compared with tabletop settings, this task poses two key challenges:
1) Spatial understanding is challenging in complex 3D environments with diverse spatial relations.
2) Action generation is difficult to generalize, as limited and costly real-robot data restricts data-driven models generalization.
To address these challenges, we propose a generalizable humanoid loco-manipulation framework that leverages the spatial perception and action generation capabilities of multi-agent large models.
Specifically, our framework includes two components: Active Spatial Brain for active spatial perception and decision-making, and Generalizable Action Cerebellum for executable robot action generation.
The first component actively perceives the spatial scene and makes decisions on task planning and subtask decomposition. 
The second component generates executable robot actions based on the decisions made by the first module without needs of task-specific real robot data.
To benchmark our framework, we design a set of spatial manipulation tasks from two perspectives: evaluating spatial perception and understanding, and assessing real-robot task performance. The results demonstrate strong performance on both aspects across diverse tasks and environments.

\end{abstract}

\section{Introduction}
\label{sec:intro}
Enabling robots to autonomously execute complex tasks in spatial environments stands as a foundational yet challenging problem in robotics and computer vision communities. Among various embodiments, humanoid robots are particularly promising due to their whole-body mobility and dexterous manipulation capabilities, which allow them to operate in complex 3D spaces. Therefore humanoid whole-body manipulation is essential for real-world applications, including home assistance and industrial scenarios.

With the development of deep learning, data-driven works have achieved remarkable success in tabletop scenarios, evolving from early parallel-jaw grasping and manipulation \cite{fang2020graspnet, wu2024economic, wang2025task, jiang2025rethinking}, to dexterous-hand grasping and manipulation \cite{xu2024dexterous, wei2024grasp, wei2025afforddexgrasp, lin2026bidexgrasp}, and more recently to tabletop manipulation on humanoid embodiments \cite{GR00t, agibot, okami, humanoidpolicy, egovla}, typically assuming fixed camera positions and constrained workspaces. Recent research extends these works to mobile manipulation platforms, including wheeled robots \cite{homerobot, aloha, momanipvla}, quadrupeds \cite{QuadWBG,human2locoman, Quad3}, and humanoids \cite{HumanoidVLA, HOMIE, Trajbooster}. Compared to tabletop settings and other robotic platforms, humanoid whole-body manipulation significantly enlarges both the observation and action spaces due to complex spatial environments and whole-body dexterity, resulting in much higher demands for large-scale, high-quality training data. Although prior efforts \cite{omniretarget, omnih2o, HMI, CLONE, clot, zhou2025mitigating, huang2026beyond} attempt to mitigate this issue by leveraging human motion capture data, simulation data, or tabletop data, as well as editing existing data to augment training for better generalization \cite{xue2025demogen, jiang2026task}, these strategies still suffer from sim-to-real gaps, embodiment discrepancies, and spatial distribution shifts. As a result, learned policies exhibit limited robustness and poor generalization in diverse and unstructured real-world environments, and even VLA models trained on diverse datasets are shown to struggle with cross-task generalization to unseen tasks \cite{zhou2026exploring}.

To address these challenges, we propose a real-robot-data-free humanoid whole-body manipulation framework by leveraging the spatial intelligence and generalizable knowledge of current vision-language models (VLMs). Fig.~\ref{fig:setting} highlights the diverse tasks our framework can accomplish. The framework comprises two coupled components: the Active Spatial Brain and the Generalizable Action Cerebellum. Together, they form a closed-loop system where the Brain decomposes the goal into manageable sub-tasks and continuously invokes proper agents in Cerebellum to solve them step-by-step.

Specifically, the Active Spatial Brain operates fundamentally as an active spatial observer and dynamic planner. To enable active spatial perception, the humanoid is equipped with a camera capable of two degrees of freedom, whose viewpoint is dynamically controlled by the VLM for active spatial understanding. We further design a memory bank to archive visual and historical context. On top of these, we introduce an adaptive task planning module that integrates closed-loop spatial reasoning with long-horizon planning, decomposing high-level instructions into specific sub-task commands.

Building on the sub-task commands produced from the brain, the Generalizable Action Cerebellum addresses the challenge of how to generate executable humanoid action in a generalizable manner. To this end, we employ several robotic action agents by leveraging the ability of foundation models without the need of task-specific robot data. Specifically, we decouple the action into lower-body and upper-body. For the lower body, two agents are provided to move the robot to where the target object lies within the dexterous hand’s workspace. For the upper body, we develop another two VLM-driven action agents that predict feasible dexterous grasp and corresponding arm motions to complete the manipulation.

To benchmark spatial-aware humanoid whole-body manipulation, we curate a set of real-world, whole-body tasks that explicitly require active spatial perception (e.g., resolving occlusions via viewpoint changes), 3D spatial reasoning (e.g., inferring relative object locations), and coordinated loco-manipulation (e.g., re-positioning the base to make targets reachable).On this benchmark,we first systematically evaluate representative VLMs in terms of complex spatial understanding. We then assess our proposed framework through real-robot experiments. The results show that our system combines reliable spatial perception and decision-making with generalizable, physically feasible whole-body action generation. Our framework achieves consistently higher success rates than existing data-driven baselines while requiring no real-robot data for training or fine-tuning.

\section{Related Work}

\subsection{Whole-Body Manipulation}
Humanoid whole-body manipulation is an active and challenging research area in embodied AI. To overcome the data bottleneck in learning-based methods, recent works utilize human teleoperation \cite{HOMIE, Twist, HMI, H2O, CLONE} for Imitation Learning \cite{WBVLA, iDP3, Being0, LeVERB}. TrajBooster \cite{Trajbooster} retargets trajectories from wheeled robots to humanoids and integrates the teleoperation data for whole-body control. ResMimic \cite{ResMimic} and $R^2S^2$ \cite{R2S2} employ task-specific residual learning or structural skill priors on top of human motion tracking. In the context of Reinforcement Learning, alongside unified controllers \cite{ULC} and force-adaptive frameworks \cite{FALCON}, AMO \cite{AMO} combines sim-to-real RL with trajectory optimization, and VIRAL \cite{VIRAL} uses large-scale visual distillation for zero-shot RL policy deployment in the real-world.
However, most methods are only applied in simplified settings, lacking active environmental interaction and reasoning by the robot. Also, they are constrained by the costly real robot data or synthetic data, making it difficult to adapt to environmental changes and limit the generalization capabilities. In contrast, our approach leverages large models for active spatial perception. By dynamically understanding 3D scenes, our multi-agent architecture autonomously decomposes tasks into executable whole-body actions, achieving superior generalization across diverse environments.

% yixiang
\subsection{LLMs-driven Manipulation}
The emergent capabilities of Large Language Models (LLMs) in sematic reasoning and zero-shot generalization \cite{LLMzeroshot, LLMfewshot, LLMCode, gptcode} have driven their extensive application in robotic manipulation. Early works in this domain primarily utilized LLMs to translate semantic instructions into structured sub-goals sequences \cite{DoasIcan, InnerMono}. To further bridge the semantic gap between planning and execution, researchers explored the synthesis of manipulation policies directly through LLMs’ code generation ability \cite{CaP, ProgPrompt}. Upon this, recent works seek to enhance physical grounding by extracting geometric representations from the tabletop workspace \cite{Voxposer, Rekep, ReSem3D, omnimanip}. These methods effectively convert language into spatial constraints and cost maps to guide the motion planning. 
While these approaches achieve promising manipulation capabilities, they are primarily designed for static tabletop scenarios with fixed camera perspectives. They inherently lack the active spatial reasoning ability and whole-body coordination mechanisms required to manage complex 3D scene-level environments and integrate locomotion and manipulation. By decoupling the problem into active spatial perception and multi-agent action generation, our system enables humanoids to dynamically reason the current scenarios and execute the whole-body action.

% junhui
\subsection{Spatial Intelligence for Robotics}
Recent advances in robotic manipulation have increasingly focused on spatial intelligence to enable robust 3D perception, spatial reasoning, and precise interaction under complex real-world conditions. Existing methods can be generally divided into three categories. The first line \cite{SpatialActor, CoPa, SOFAR} focuses on explicit spatial modeling and geometric-semantic disentanglement, which decouples and fuses geometric and semantic information to infer actionable 3D structures and poses for accurate manipulation. The second \cite{internvlam1, ActiveVLA, SpatialPolicy} adopts spatial-aware vision-language-action (VLA) frameworks, unifying spatial grounding, language understanding, and action generation in an end-to-end pipeline to improve generalization via embedded spatial reasoning. The third\cite{RoboPoint, MOKA, AimBot} exploits spatial affordance and keypoint prediction, often using lightweight augmentation or vision-language model fine-tuning to localize operable regions with low computational overhead. Although these efforts lay the foundation for advanced embodied intelligence, they are mostly limited to tabletop scenarios or simple loco-manipulation tasks. In contrast, this work targets spatial-aware humanoid whole-body manipulation, which integrates active spatial perception and complex decision-making by fully exploiting the strong spatial reasoning and decision-making capabilities of multimodal large language models.

\begin{figure*}[tbp]
  \centering
  \includegraphics[width=\textwidth]{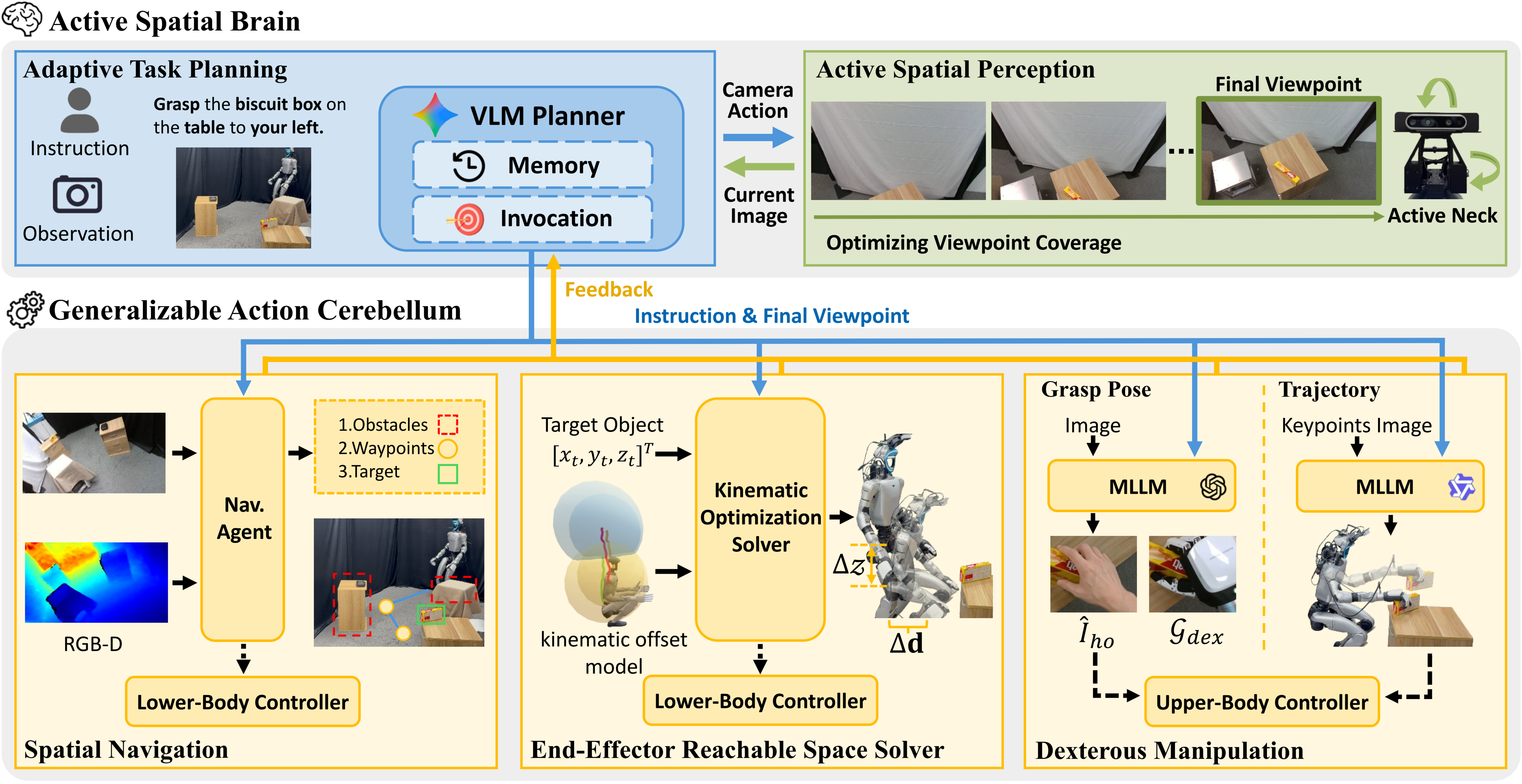}
  \caption{The overview of our humanoid whole-body manipulation framework. Our framework consists of two components: an Active Spatial Brain for active spatial perception, understanding and planing; and a Generalizable Action Cerebellum for executable action generation.}
  \label{fig:main}
\end{figure*}

% yilin
\section{Methods}
\subsection{Framework Overview}
\subsubsection{Problem Formulation}
In this paper, we focus on the humanoid whole-body manipulation task. Given user language commands, RGB-D observations, and the proprioceptive state, the framework aims to generate the whole-body action of the humanoid, denoted as $\mathcal{A} = \{a_{cam}, a_{dex}, a_{upper}, a_{lower}\}$, each referring to the active camera action $a_{cam}$, dexterous hand poses $a_{dex}$, upper-body joint poses $a_{upper}$, and lower-body joint poses $a_{lower}$. Finally, $\mathcal{A}$ is sent to PD controllers to output joint torques.

\subsubsection{Framework Overview}

Our framework adopts a hierarchical multi-agent architecture for humanoid whole-body manipulation in complex environments, as illustrated in Fig.~\ref{fig:main}. It consists of two tightly coupled components: an Active Spatial Brain and a Generalizable Action Cerebellum. Operating as a high-level policy, the Brain actively explores the spatial environment, decomposes the goal into sub-tasks, and invokes specific action agents within the Cerebellum to execute low-level robot actions to complete them. Once action agents are invoked, feedback is routed back to the Brain. Together, these two components work in a perception-planning-control closed-loop until the goal is successfully completed.

\subsection{Active Spatial Brain}
To enable the humanoid robot to autonomously perceive complex 3D environments, understand natural language instructions and make manipulation planing, we propose the Active Spatial Brain empowered by the mainstream VLM. The Brain integrates three modules: Active Spatial Perception for active surrounding perception, Memory Bank archiving perceived observations to support spatial awareness consistency, and Adaptive Task Planning leveraging them to generate spatially adaptive task plans.

\begin{figure*}[tbp]
  \centering
  \includegraphics[width=0.65\textwidth]{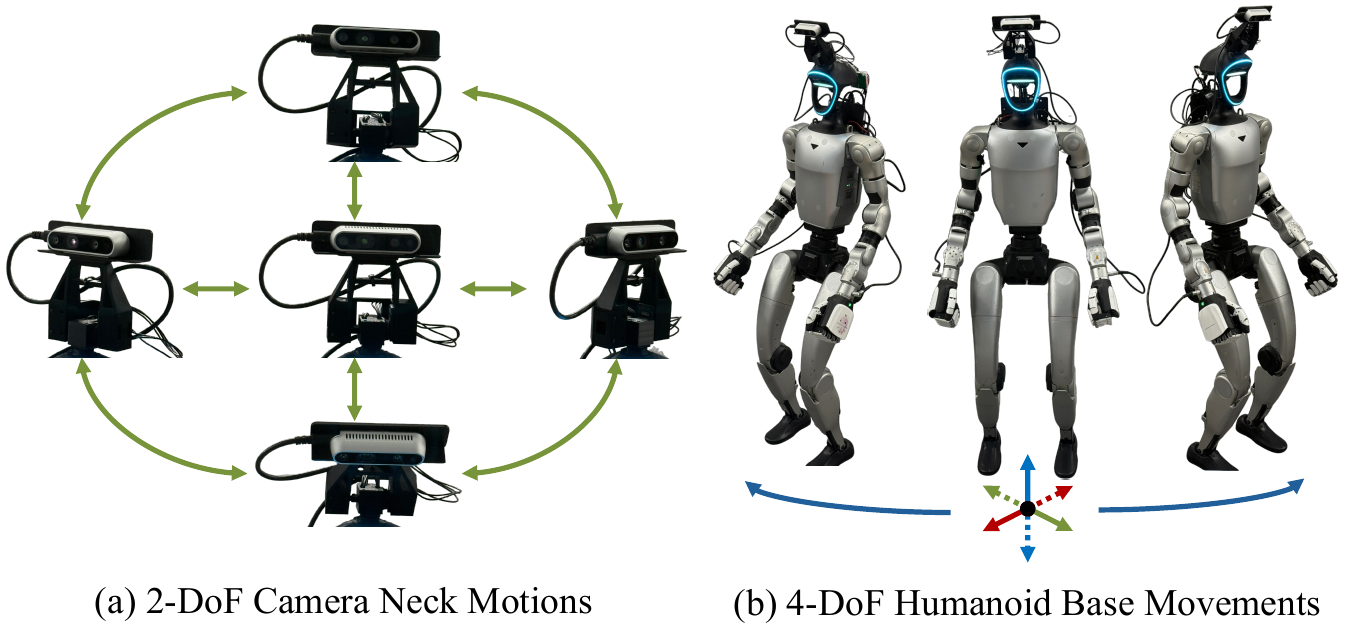}
  \caption{The illustration of the degrees of freedom in the active camera, consisting of two parts: 2-DoF camera neck motions and 4-DoF camera base changes induced by humanoid body movements.}
  \label{fig:camera}
\end{figure*}

\subsubsection{Active Spatial Perception}
In complex real-world scenes, objects are frequently occluded or arranged in intricate spatial configurations, making fixed-perspective perception inadequate for accurate localization and manipulation. To address this challenge, we propose an active spatial perception agent to dynamically adjust the camera viewpoint to improve spatial understanding in unstructured 3D environments. We fully leverage the spatial reasoning capability of VLM to make viewpoint decisions directly from visual observations and task instructions. Specifically, we prompt the VLM with a pre-defined robot action space and its corresponding effects, where the space consists of a 2-DoF camera neck motion and a 4-DoF humanoid base movement including translations along the $x$, $y$, and $z$ axes and $z$-axis rotation (Fig.~\ref{fig:camera}). With this capability, the agent can actively explore complex environments to locate task-relevant targets.

\subsubsection{Memory Bank}
 In loco-manipulation, drastic shifts in the robot’s base position can cause inherent viewpoint inconsistencies that mislead the VLM. To address this, we implement a hybrid recording movement-conditioned strategy. Specifically, the bank archives only the images captured by the perception module since the last base movement, pairing each with its corresponding camera pose. Coupled with constrained visuals, it maintains a textual log recording all commands dispatched from the Brain to the Cerebellum, alongside their respective feedback. By synthesizing these selectively archived images and the global text-based execution history, the framework enables the VLM to accurately infer the task stage and maintain consistent spatial awareness.

\begin{figure}[tbp]
    \centering
    \includegraphics[width=0.8\linewidth]{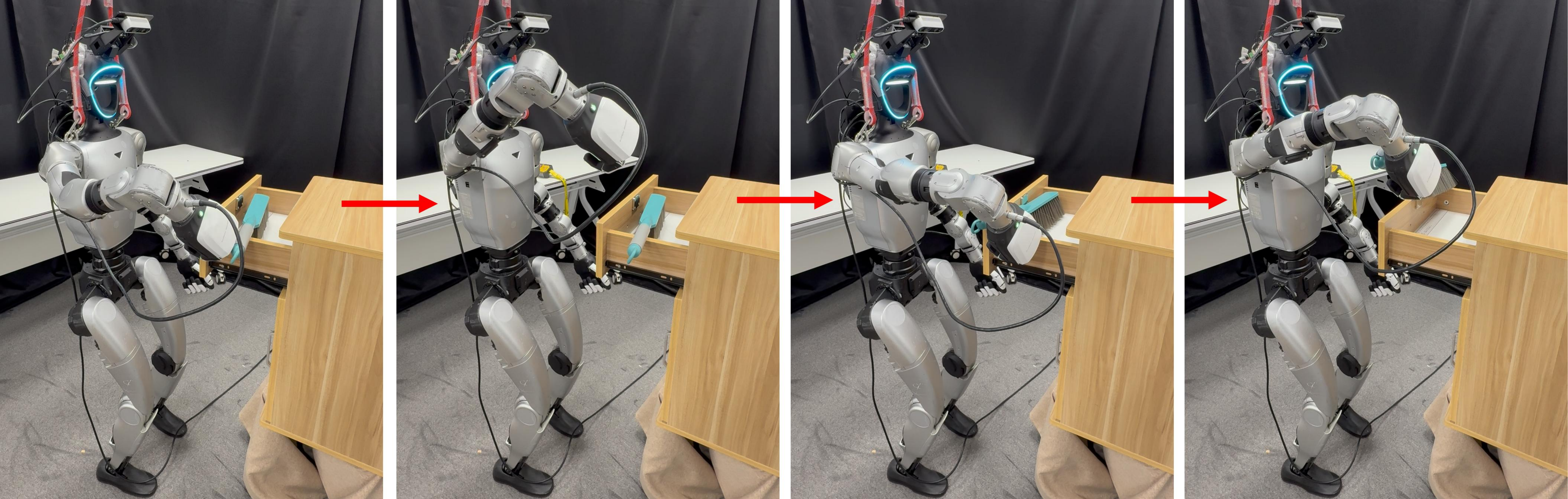}
    \caption{The planner adjusts the plan via execution history and visual validation. Left to right: the robot misses the target, readjusts its pose, and successfully grasps it.}
    \label{fig:retry}
\end{figure}

\subsubsection{Adaptive Task Planning}
Building upon the active spatial perception module and the memory bank, the adaptive task planning agent integrates spatial reasoning, sub-task decomposition, and dynamic replanning. Given a user instruction, the agent first decomposes the long-horizon goal into a sequence of sub-tasks, followed by an iterative closed-loop execution. At each step, the planner evaluates the sequence against the memory bank to determine the current task stage and triggers one of two behaviors: (i) active viewpoint adjustment to achieve more comprehensive spatial perception, or (ii) invoking downstream agents in the Cerebellum to progress the overall plan. 

However, the initial plan may be partially infeasible due to insufficient observations, and environmental disturbances can result in agent execution failures. To mitigate these issues, the planner dynamically replans its sub-tasks based on the memory bank at each step. For instance, upon detecting a failed grasp, it autonomously re-invokes the manipulation agent to execute a second attempt (Fig. \ref{fig:retry}). This unified replanning process achieves robust closed-loop control in complex, unstructured 3D environments.

% linmu
\subsection{Generalizable Action Cerebellum}
To translate high-level commands from the Spatial Brain into executable and physically feasible humanoid action sequences in a generalizable manner, we introduce the Generalizable Action Cerebellum, which decomposes whole-body control into lower-body and upper-body control, executed by multiple specialized action agents. The lower-body locomotion module handles obstacle-aware navigation and base adjustments to ensure end-effector reachability. The upper-body dexterous manipulation module generates physically feasible upper-body motions for manipulating the objects. By coordinating specialized action agents for both components without requiring task-specific demonstrations, the system achieves robust and generalizable whole-body humanoid manipulation.

\subsubsection{3.3.1 Humanoid Lower-Body Locomotion}~
\newline

\noindent To achieve reliable and precise robot locomotion for subsequent manipulation, the lower-body locomotion module is built around two agents that can be invoked by the Brain based on perceived spatial relationships. In terms of their roles, a navigation agent manages long-range obstacle-aware coarse approach, while a reachable space solving agent provides short-range positioning to secure the object within the manipulation workspace. Beneath them, an RL policy is trained to convert the base movements from the agents into executable joint actions.

\subsubsection{Spatial Navigation}
For obstacle-aware navigation, this agent requests an RGB-D observation which manifests the target along with the ground, and the 2D image coordinate of the target $\mathbf{u}_{\text{target}}=(u_{tar},v_{tar})$. We first lift the depth map into a 3D point set in the camera frame. The resulting point cloud is then projected onto the ground plane to construct a discretized grid map with cell size $d$.

A grid cell is marked as occupied if there exists a point within the cell whose height exceeds a threshold $h$:
\begin{equation}
\mathcal{G}(i,j)=
\begin{cases}
1, & \exists\,\mathbf{x}\in \mathcal{C}_{ij} \ \text{s.t.}\ z(\mathbf{x})>h,\\
0, & \text{otherwise}.
\end{cases}
\end{equation}

The target coordinate $\mathbf{u}_{\text{target}}$ undergoes the same lifting and projection process to determine the corresponding goal cell in the grid map.

After constructing the grid map, we apply the A* algorithm to compute a collision-free path from the robot origin to the goal cell. The resulting path is then simplified using the Ramer–Douglas–Peucker (RDP) algorithm, producing a waypoint sequence $\{\mathbf{p}_i\}_{i=1}^{N}$ that serves as the final trajectory. The robot tracks $\mathbf{p}_i$ using closed-loop feedback, computing $v_{x,t}$ and $\omega_{y,t}$ from the relative pose to the next waypoint while maintaining a nominal base height $h_t$.

\subsubsection{End-Effector Reachable Space Solver}

To ensure the EE's reachability at close range, the robot should make precise adjustments to its base position. We therefore introduce an optimization based solver that computes the required base adjustment.

Given the target object position in the original robot's base frame $\mathbf{p}_{t} = [x_t, y_t, z_t]^T$, let $\mathbf{p}_{\text{obj}}( \mathbf{p}_{t}, \Delta \mathbf{d}, \Delta z)$ denote the target object position expressed in the adjusted base frame, where $\Delta \mathbf{d}\in\mathbb{R}^2$ and $\Delta z$ represent the horizontal and vertical base adjustments. The solver minimizes the adjustment magnitude while bringing the object closer to the shoulder workspace:
\begin{equation}
\min_{\Delta\mathbf{d},\, \Delta z}
\;\lambda_1 \|\Delta \mathbf{d}\|_2^2
+ \lambda_2 (\Delta z)^2
+ \lambda_3 
\left\|
\mathbf{p}_{\text{obj}}( \mathbf{p}_{t}, \Delta \mathbf{d}, \Delta z)
-
\mathbf{p}_{\text{shoulder}}(\Delta z)
\right\|_2^2 .
\end{equation}

To ensure reachability, we impose a constraint on the shoulder-to-object distance:
\begin{equation}
\left\|
\mathbf{p}_{\text{obj}}( \mathbf{p}_{t}, \Delta \mathbf{d}, \Delta z)
-
\mathbf{p}_{\text{shoulder}}(\Delta z)
\right\|_2
\le \eta L,
\end{equation}
where $L$ is the arm length and $\eta=0.8$ defines a safety margin.

Changing the base height alters the relative pose between the base and the shoulder due to the humanoid kinematic structure. To account for this effect, we estimate a kinematic offset model $\Phi(z)$ using MuJoCo\cite{todorov2012mujoco} simulation:
\begin{equation}
\Phi(z) = (\delta x(z),\, \delta z(z)),
\end{equation}
which predicts the horizontal and vertical offsets of the shoulder relative to the base at height $z$.

The shoulder position in the base frame is therefore

\begin{equation}
\mathbf{p}_{\text{shoulder}}(\Delta z)
=[\delta x(z_0+\Delta z), \ 0, \ \delta z(z_0+\Delta z)]^T
\end{equation}

where $z_0$ is the nominal base height. Solving this optimization yields the refined base displacement and height that place the object within a safe and reachable region for subsequent manipulation. We then compute the $v_{x,t}$, $\omega_{y,t}$, $h_t$ corresponding to $\Delta d$ and $\Delta z$ for the RL policy.

\subsubsection{Reinforcement Learning-Based Humanoid Locomotion}

We finally adopt a policy $\pi_{\text{loco}}$ following HOMIE\cite{HOMIE} to translate the base commands from agents above to actual lower-body joints actions. Given the proprioception state $\mathbf{s}_t$ and base command $\mathbf{c}_t = [v_{x,t}, \omega_{y,t}, h_t]$, the policy outputs $\mathbf{a}_{\text{lower},t} = \pi_{\text{loco}}(\mathbf{s}_t, \mathbf{c}_t)$ to achieve stable lower-body locomotion.
% 在得到了

\subsubsection{3.3.2 Humanoid Upper-Body Dexterous Manipulation}~
\newline

 \noindent Humanoid upper-body dexterous manipulation requires precise control of both arm and dexterous hand joints, posing challenges due to the high degrees of freedom and limited generalization across tasks and objects. To address these challenges, we decompose manipulation into grasp generation and post-grasp trajectory generation implemented by two agents empowered by MLLM. One synthesizes object-specific grasp poses, while the other one produces manipulation trajectories after grasp acquisition.

\subsubsection{Grasp Pose Generation} Inspired by the paradigm of leveraging foundation models to generate human grasp images and transferring them into executable robot actions~\cite{wei2026omnidexgrasp}, the grasp pose generation agent leverages the multimodal foundation generative models to synthesize hand--object interaction images $\hat{I}_{ho}$ conditioned on the visual observation $I$ and task instruction $\mathcal{L}$. By harnessing the foundation model's ability to learn generalizable grasp patterns from large-scale data, these images encode transferable human grasp strategies that serve as a strong prior for subsequent robot grasp generation, even though they cannot be executed directly due to embodiment and physical constraints.

To bridge this gap, the agent performs a human-image-to-robot grasp transfer via hand--object reconstruction and dexterous retargeting. In the hand--object reconstruction stage, the agent first invokes the pretrained Hyper3D\cite{hyper3d2024} model to generate a 3D object mesh $\mathcal{M}_o$ from the observation $I$. Given $\mathcal{M}_o$ and $\hat{I}_{ho}$, the agent then applies EasyHOI\cite{liu2025easyhoi} to estimate the human hand pose $\mathcal{G}_{mano}$. The reconstructed human grasp is then retargeted to the dexterous hand by aligning the fingertip positions $p^{dex,ft}_k$ with the corresponding human fingertip positions $p^{mano,ft}_k$, and the optimization objective is formulated as:
\begin{equation}
\min_{\mathcal{G}_{dex}} \sum_k \left\| p^{dex,ft}_k - p^{mano,ft}_k \right\|_2^2 .
\end{equation}
The optimized $p^{dex,ft}_k$ are then retargeted to dexterous hand actions $\mathcal{G}_{dex}$ as the hand pose output.

\begin{figure}[t]
    \centering
    % 第一行
    % (a) Push
    \begin{subfigure}{0.35\linewidth}
        \centering
        \includegraphics[width=\linewidth, trim=17mm 5mm 17mm 5mm, clip]{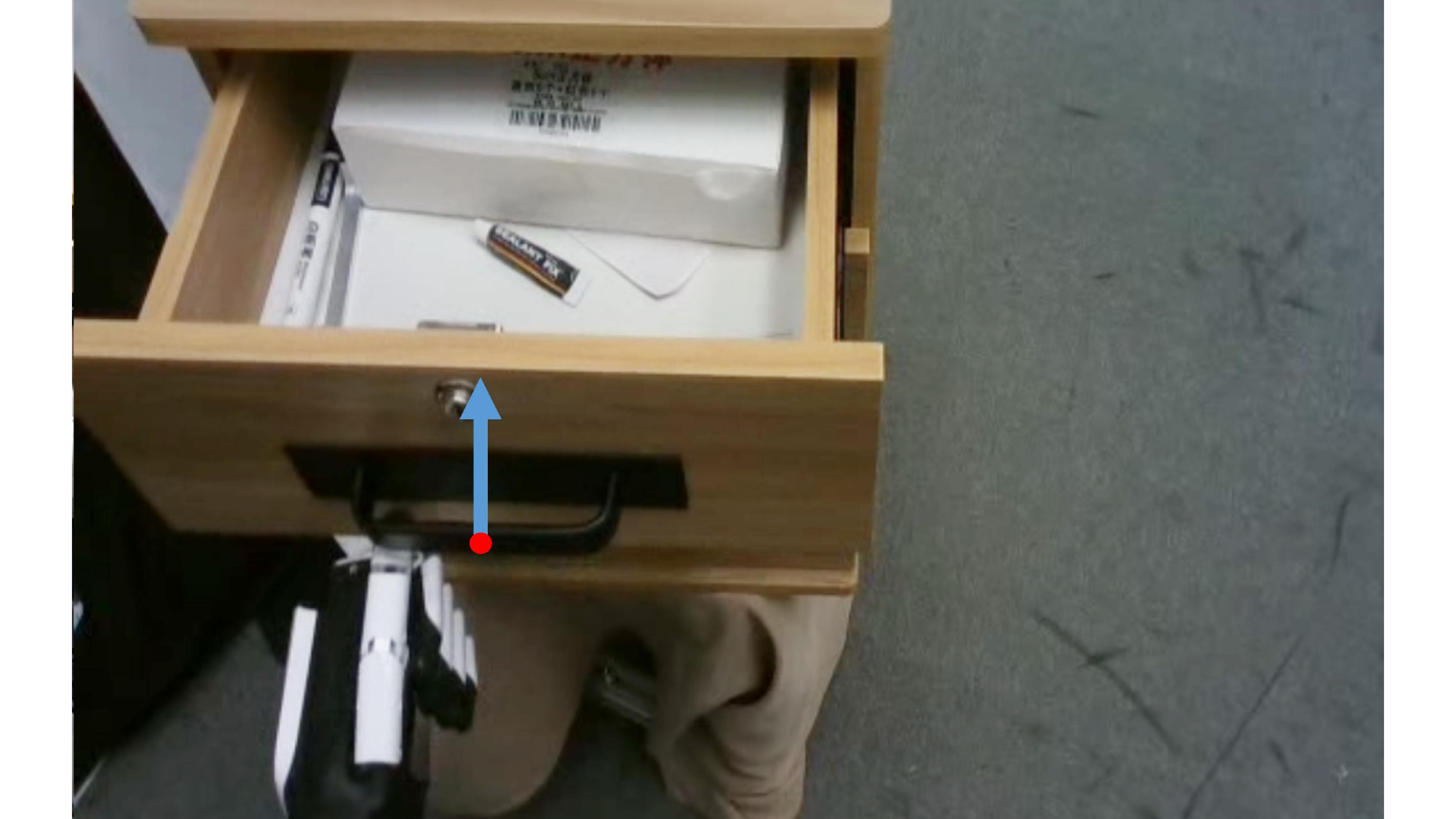}
        \caption{Push}
        \label{fig:prim_push}
    \end{subfigure}
    % (b) Pull
    \begin{subfigure}{0.35\linewidth}
        \centering
        \includegraphics[width=\linewidth, trim=17mm 5mm 17mm 5mm, clip]{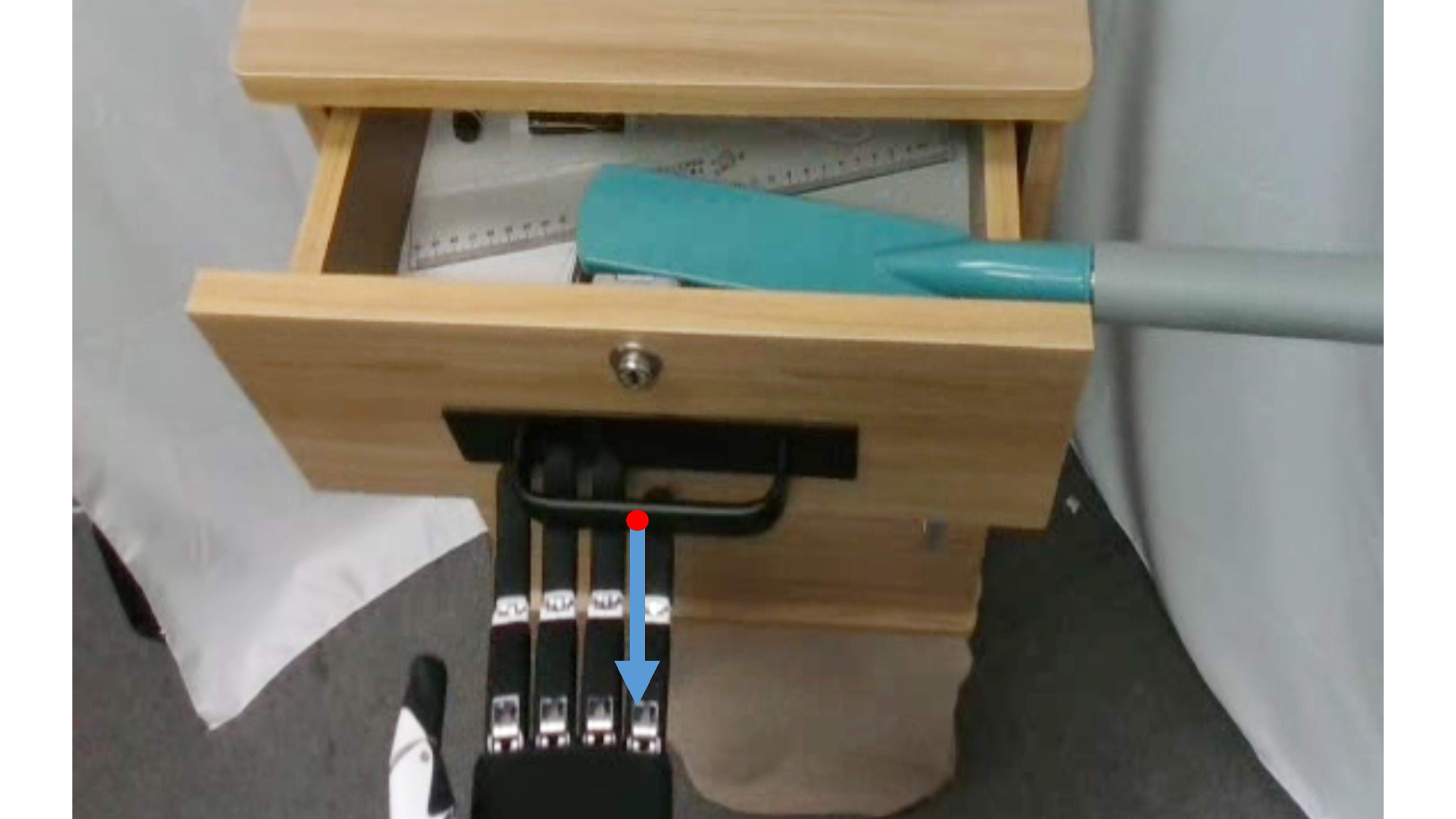}
        \caption{Pull}
        \label{fig:prim_pull}
    \end{subfigure}

    % 第二行
    % (c) Place
    \begin{subfigure}{0.35\linewidth}
        \centering
        \includegraphics[width=\linewidth, trim=17mm 5mm 17mm 5mm, clip]{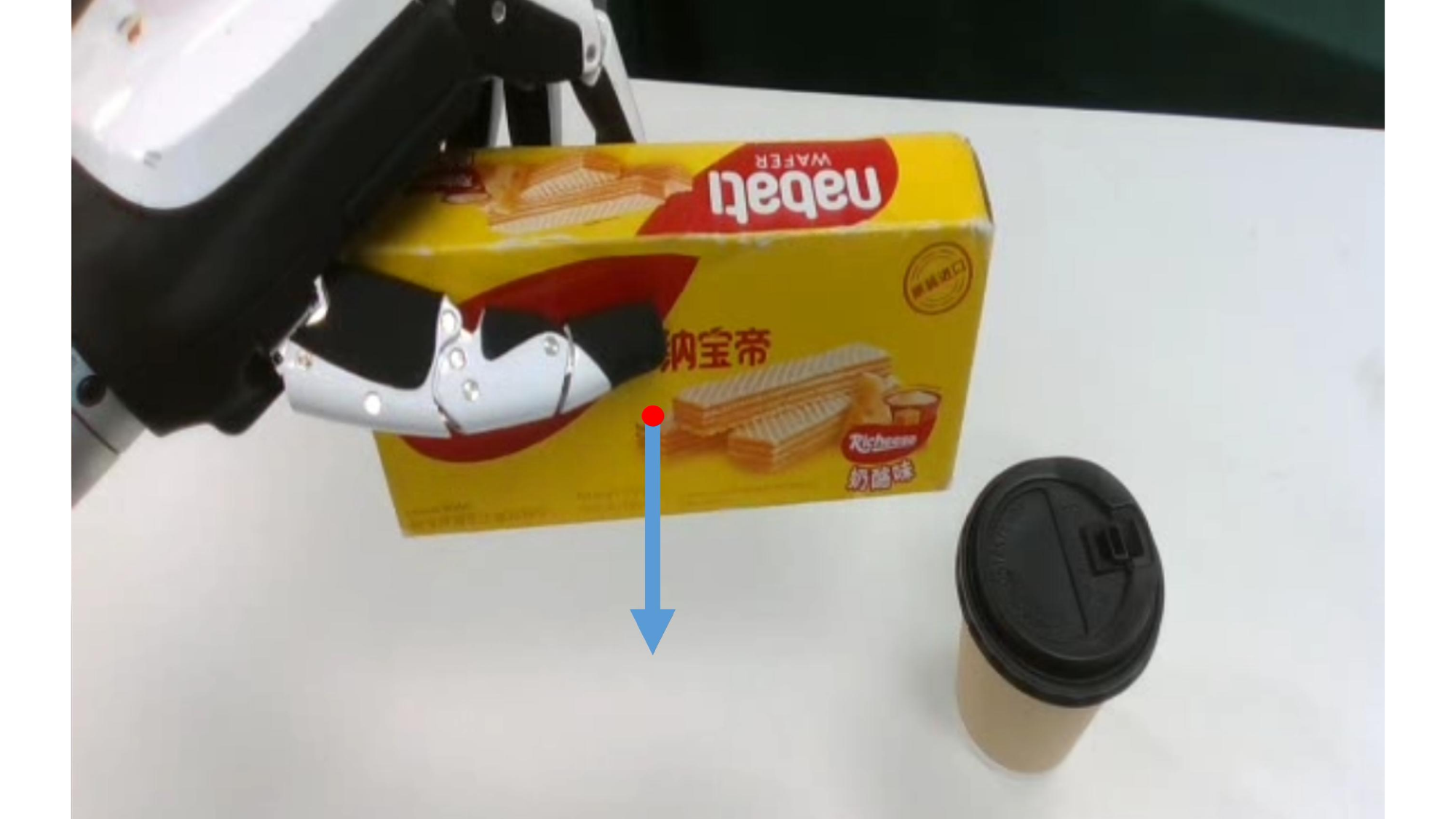}
        \caption{Place}
        \label{fig:prim_place}
    \end{subfigure}
    % (d) Rotate
    \begin{subfigure}{0.35\linewidth}
        \centering
        \includegraphics[width=\linewidth, trim=17mm 5mm 17mm 5mm, clip]{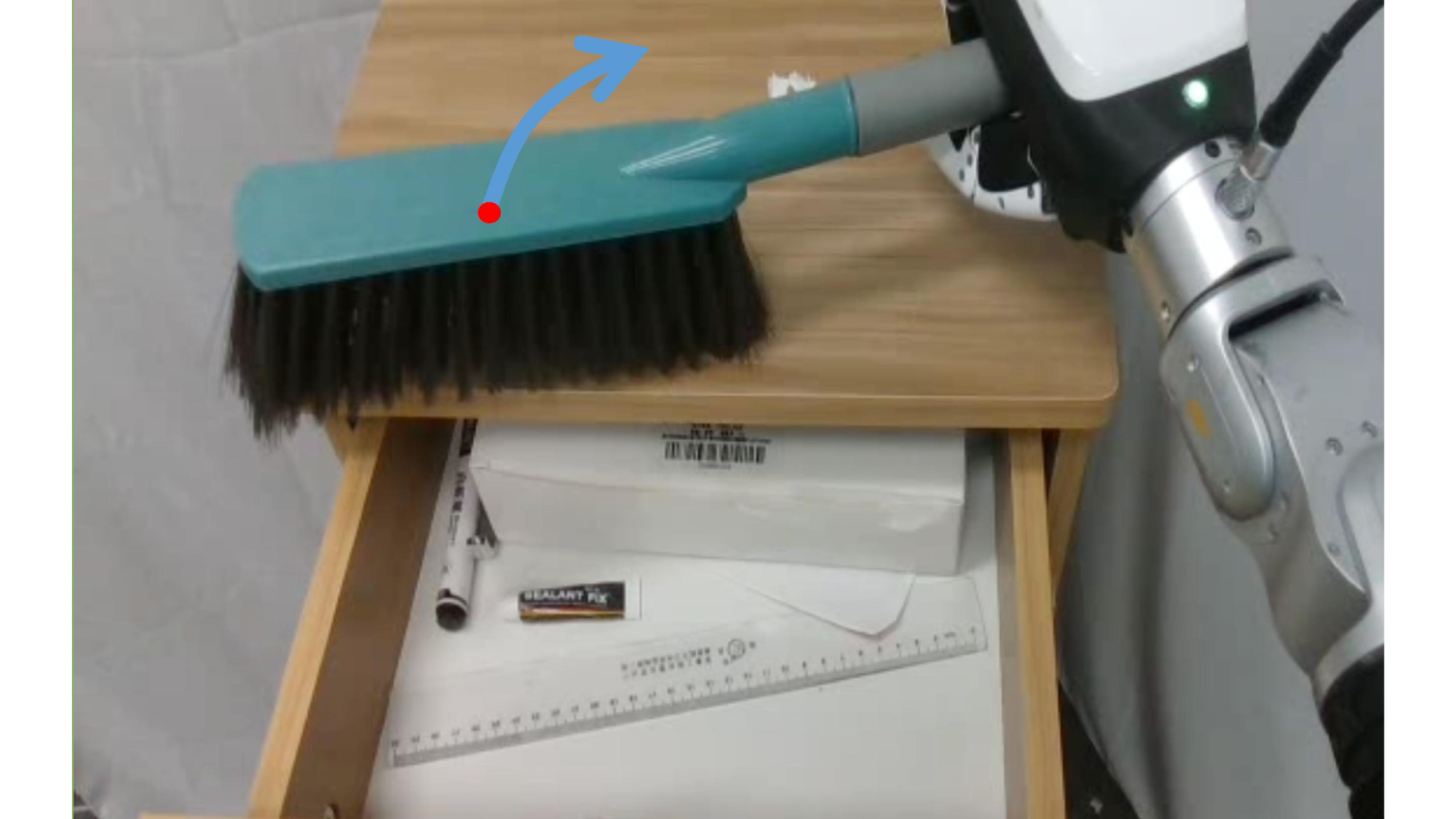}
        \caption{Rotate}
        \label{fig:prim_rotate}
    \end{subfigure}
    
    % 精简后的 Caption
    \caption{Implementation of fundamental manipulation primitives. Red dots denote target spatial keypoints, and blue arrows indicate trajectory directions.}
    \label{fig:primitives}
\end{figure}

\subsubsection{Post-grasp Trajectory Generation} 
This agent identifies the post-grasp trajectory through parameterized action primitives (Fig.~\ref{fig:primitives}). Given a visual observation and language instruction, the large model predicts semantic 2D keypoints and selects a corresponding action primitive. After lifting these keypoints to 3D coordinates, we derive the target 6-DoF EE poses by applying the primitive's specific kinematic rules. All primitives begin with an alignment phase, moving the EE to a target keypoint or preparatory pose. Subsequently, \textit{Pushing} (Fig.~\ref{fig:prim_push}) and \textit{Pulling} (Fig.~\ref{fig:prim_pull}) apply linear translation to manipulate articulated objects like drawers. \textit{Placing} (Fig.~\ref{fig:prim_place}) preserves the initial hand-object spatial offset at the new location, while \textit{Rotating} (Fig.~\ref{fig:prim_rotate}) executes a pivot-based local rotation. Finally, an IK solver converts these sequential poses into continuous joint commands for whole-body execution.

\begin{figure}[t]
  \centering
  \includegraphics[width=0.6\textwidth]{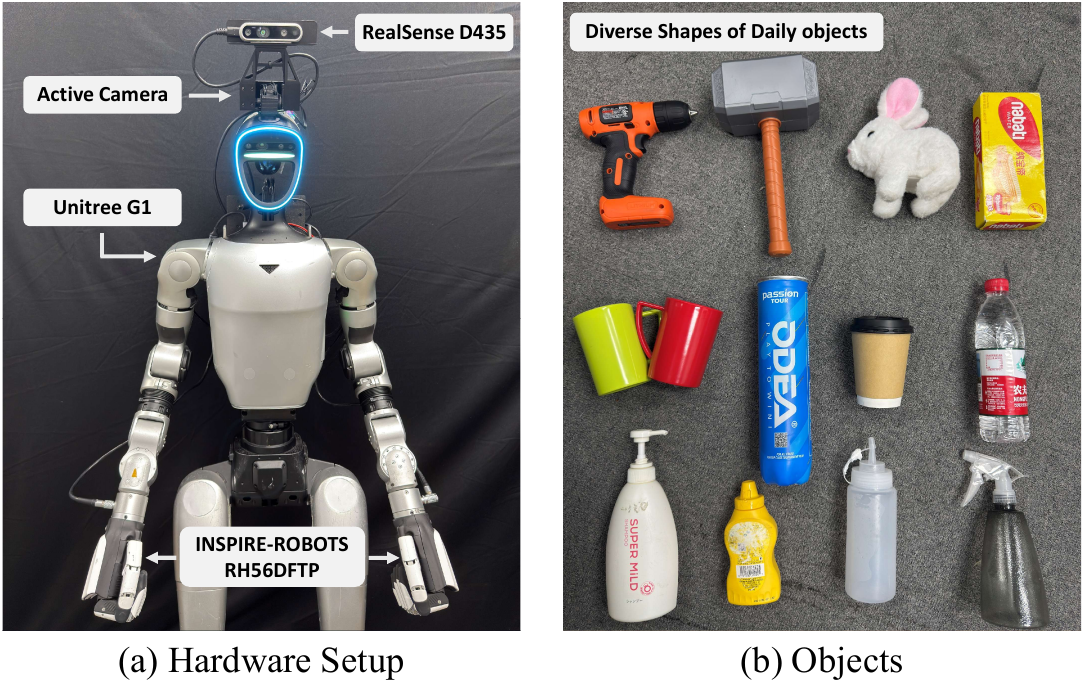}
  % \caption{The hardware setup and objects used in our experiments.}
  \caption{Real-world experimental setup. (a) The robot hardware platform. (b) The objects to be grasped in our experiments.}
  \label{fig:real_setting}
\end{figure}
\section{Experiments}
%supp的benchmark
\subsection{Spatial Intelligence Benchmark for Manipulation Tasks}

Spatial intelligence constitutes one of the cores of our framework. To investigate the spatial intelligence capabilities of mainstream VLMs, we systematically evaluate them through several tasks. These tasks assess spatial relationship understanding, 3D scene modeling, and active perception, respectively.

\textbf{Task 1} Map a shuffled set of nine adjacent images back to their original $3 \times 3$ spatial grid, where proficiency is quantified by structural matching accuracy.

\textbf{Task 2} The VLM drives the active camera to locate an initially out-of-view or occluded target, continuing until successful detection or running out budget.

\textbf{Task 3}  Predict a sequence of ground waypoints leading to the target within an obstacle-cluttered scene. Trajectory formed by waypoints is graded by obstacle clearance as Appropriate (moderate distance), Inappropriate (excessively close/far) or Fail (collision).

\subsection{Real World Experiments Setting}
\subsubsection{Robot Platform}

% We conduct real world experiments on a Unitree G1 humanoids with two 6-DOF Inspire FTP Hand and a 2-DOF active Realsense D435i camera.
We conduct real world experiments on a Unitree G1 humanoid equipped with two 6-DoF Inspire FTP hands and a 2-DoF active Realsense D435i camera (Fig.~\ref{fig:real_setting}(a)). The set of objects used for manipulation is shown in Fig.~\ref{fig:real_setting}(b).

\subsubsection{Task Description}
To evaluate the proposed humanoid mobile manipulation framework under diverse conditions, we design five task settings. Each setting includes an \textit{easy} and a \textit{hard} mode.

1. \textbf{Different Heights.} The robot performs manipulation without base translation while adjusting its body height. The easy mode uses moderate heights, while the hard mode includes more extreme height variations.

2. \textbf{Different Positions.} The robot maintains a fixed height but navigates to the target without obstacles. The easy mode requires short-range motion, whereas the hard mode involves longer distances.

3. \textbf{Different Heights and Positions.} The robot must adjust both height and position. The easy mode corresponds to the intersection of the easy cases above, and the hard mode to the union of their hard cases.

4. \textbf{Obstacle Avoidance.} The robot avoids obstacles during navigation, with one obstacle in the easy mode and multiple obstacles in the hard mode.

5. \textbf{Occlusion Handling.} The robot removes occluding objects before manipulating the target. The easy mode uses a fixed occluder, while the hard mode includes diverse occluding objects.

For each setting, we conduct 10–30 trials and report the success rate, ensuring that the number of trials is identical across methods for fair comparison. A trial is considered successful only if the robot completes the manipulation task without any collision with obstacles.

\subsection{Can our framework surpass data-driven methods?}
\subsubsection{Reproduction of Data-driven Method.} 
We select TrajBooster\cite{Trajbooster} and $\Psi_0$\cite{psi0} as the learning-based baseline for comparison. We initialize the model with the officially released post-pre-trained weights and fine-tune it on our task settings. For each setting, we collect at least 20 trajectories under the easy configuration for post-training. The fine-tuned model is then evaluated on both the easy (in-domain) and hard (out-of-domain) configurations to assess its generalization performance.

\begin{table}[t]
  \centering
  \caption{Comparison with data-driven (TrajBooster, $\Psi_0$) and VLM-driven (CaP) baselines on diverse tasks. Easy denotes in-domain tasks, while Hard denotes more complex out-of-distribution tasks. }  
  \label{tab:task_performance}
  \footnotesize % 保持字体较小，更显专业
    \renewcommand{\arraystretch}{1.0}
    \setlength{\tabcolsep}{12.2pt} % 调节列间距
    \setlength{\aboverulesep}{0pt}
    \setlength{\belowrulesep}{0pt}

  \begin{tabularx}{\textwidth}{lccccc} % X 代表自动伸缩且换行的列
    \toprule
    \textbf{Method} & \textbf{Task 1} & \textbf{Task 2} & \textbf{Task 3} & \textbf{Task 4} & \textbf{Task 5} \\
    \bottomrule
    \rowcolor{gray!20}
    \multicolumn{6}{l}{\textit{\textbf{Easy Setting}}} \\
    \midrule
    TrajBooster & 70.0 & \textbf{75.0} & 55.0 & 60.0 & 40.0 \\
    $\Psi_0$    & 80.0   & \textbf{75.0}   & 70.0 & 70.0 & 60.0 \\
    CaP         & 75.0   &  70.0  & 70.0 & \textbf{80.0} & 65.0 \\
    Ours        & \textbf{85.0} & \textbf{75.0} & \textbf{80.0} & \textbf{80.0} & \textbf{70.0} \\
    \bottomrule
    \rowcolor{gray!20}
    \multicolumn{6}{l}{\textit{\textbf{Hard Setting}}} \\
    \midrule
    TrajBooster & 20.0 & 40.0 & 10.0 & 0    & 20.0 \\
    $\Psi_0$    & 35.0   & 45.0   & 25.0 & 0    & 30.0 \\
    CaP         & 50.0   & 55.0   & 35.0 & 30.0 & 35.0 \\
    Ours        & \textbf{60.0} & \textbf{70.0} & \textbf{55.0} & \textbf{60.0} & \textbf{60.0} \\
    \bottomrule
  \end{tabularx}
\end{table}

% 没改，单纯复现，数据只包括easy的数据，没有包括hard的数据
\subsubsection{Comparison Results.}
As shown in Table~\ref{tab:task_performance}, our framework consistently outperforms both data-driven baselines (TrajBooster, $\Psi_0$) and the VLM-driven baseline (CaP\cite{CaP}) on spatial humanoid manipulation tasks. To analyze performance under different difficulty levels, we divide the tasks into easy and hard settings. The easy setting involves simple spatial relationships and limited generalization demands, whereas the hard setting requires stronger spatial reasoning and broader generalization in complex environments. We observe that data-driven methods remain competitive on the easy tasks, but their performance drops significantly on the hard tasks. This is likely because such methods depend heavily on the quality and diversity of training data, making them less robust to out-of-distribution scenarios and complex spatial relationships not covered during training. Additionally, existing VLM methods (e.g., CaP) generalize well but underperform on humanoids without specific designs. In contrast, our framework maintains strong performance in both settings, demonstrating better generalization through the multi-agent design.

\begin{figure}[t]
\centering
\includegraphics[width=\linewidth]{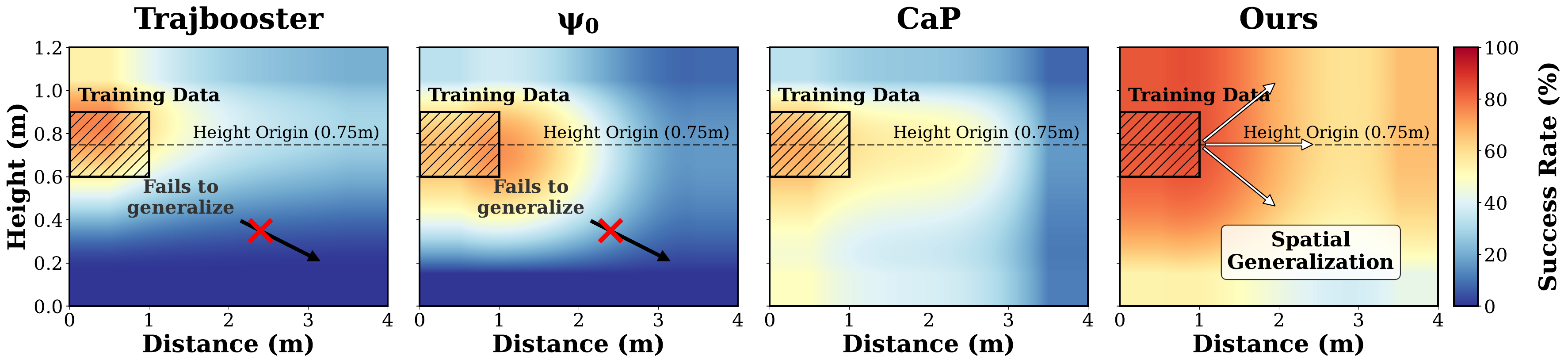}
\caption{Spatial reachability heatmaps across methods. The heatmap visualizes the reachable regions under different target locations, showing that our framework maintains more robust reachability than the data-driven baseline.}
\label{fig:reachability_heatmap}
\end{figure}

\subsection{Can our framework achieve generalization?}
\begin{wraptable}{r}{0.46\linewidth}
\centering
\scriptsize
\setlength{\tabcolsep}{4pt}
\renewcommand{\arraystretch}{1.05}
\caption{Generalization across object categories. * indicates objects unseen in the fine-tuning dataset of TrajBooster.}
\label{tab:object_generalization}
\begin{tabular}{>{\centering\arraybackslash}m{1.0cm}
                >{\centering\arraybackslash}m{0.7cm}
                >{\centering\arraybackslash}m{0.9cm}}
\toprule
\textbf{Objects} & \textbf{Ours} & \textbf{Traj.} \\
\midrule
Tools  & \textbf{77.8} & 50.0 \\
Tools* & \textbf{52.0} & 50.0 \\
Items  & \textbf{60.0} & \textbf{60.0} \\
Items* & \textbf{69.6} & 50.0 \\
\bottomrule
\end{tabular}
\end{wraptable}

We further analyze the generalization capability of our framework from two complementary perspectives: spatial reachability and object categories. Fig.~\ref{fig:reachability_heatmap} visualizes the reachable regions across different target locations, showing that our framework maintains a broader and more reliable reachable region than the data-driven baseline. Table~\ref{tab:object_generalization} further demonstrates strong generalization across both seen and unseen object categories. In contrast, data-driven methods perform reasonably well when the test scenarios are similar to the training domain, but their performance degrades significantly when encountering unseen spatial layouts or novel objects. These results highlight the advantage of our approach in handling both reachability variations in space and category variations in real-world environments.

\subsection{Is each component of our framework effective?}
We conduct an ablation study to evaluate the key modules: Active Perception (AP) and End-Effector Reachable Space Solver (EES). In detail, Robot's viewpoint is fixed without AP, and it maintains a predefined distance and relative height difference with the target object without EES. Shown in Table ~\ref{tab:ablation_study}, Task 3 performance is consistently low without AP, confirming that AP significantly expands the robot’s observation space to maintain visual context during long-distance base movements. Success rates all drop without EES in the experiment, especially in Task 4 where sensor errors accumulated during non-linear movement, showing that EES further improves manipulation by refining object alignment and positioning. Results demonstrate that our design effectively improves humanoid ability in whole-body manipulation tasks.

\begin{table}[h]
    \centering
    \caption{Ablation study of our framework on two representative tasks.}
    \label{tab:ablation_study}
    % 如果觉得左右边缘还是太挤，可以把 2pt 改成 4pt，或者直接删掉下面这行
    \setlength{\tabcolsep}{4pt} 
  \scriptsize 
    \begin{tabular*}{1.0\columnwidth}{@{\extracolsep{\fill}} cccc }
        \toprule
        \textbf{AP} & \textbf{EES} & \textbf{Diff. Heights and Positions} & \textbf{Obstacle Avoidance} \\
        \midrule
        $\times$ & $\times$ & 18.8 &  0\\
        $\checkmark$ & $\times$ & 55.6 & 16.7 \\
        $\times$ & $\checkmark$ & 24.9 & 33.4 \\
        \midrule
        $\checkmark$ & $\checkmark$ & \textbf{67.5} & \textbf{70.0} \\
        \bottomrule
    \end{tabular*}
\end{table}

\begin{table}[t]
\centering
\scriptsize
\setlength{\tabcolsep}{1pt} 

\caption{Spatial intelligence benchmark of VLMs. Time: average seconds per task.}
\label{tab:benchmark}

% ================= (a) 调整为分配 38% 宽度 =================
\begin{subtable}[t]{0.42\linewidth}
\centering
\begin{tabular*}{\linewidth}{@{\extracolsep{\fill}}lcc@{}}
\toprule
\textbf{Models} & \textbf{Correctness} & \textbf{Time} \\
\midrule
gpt5.1 & 94.12 & 94.59 \\
gemini-3-flash & 95.21 & \textbf{54.93} \\
gemini-3.1-pro & \textbf{98.84} & 94.50 \\
qwen3.5-plus & 93.29 & 155.63 \\
\bottomrule
\end{tabular*}
\caption{Map images back to grid}
\label{tab:sub_a}
\end{subtable}%
\hfill % 自动填充中间的缝隙
% ================= (b) 调整为分配 58% 宽度 =================
\begin{subtable}[t]{0.54\linewidth}
\centering
\begin{tabular*}{\linewidth}{@{\extracolsep{\fill}}lccc@{}}
\toprule
\textbf{Models} & \textbf{Success} & \textbf{Observe Times} & \textbf{Time} \\
\midrule
gpt5.1 & 82.35 & 2.53 & 45.75 \\
gemini-3-flash & 88.23 & \textbf{2.00} & \textbf{27.07} \\
gemini-3.1-pro & \textbf{94.11} & \textbf{2.00} & 44.73 \\
qwen3.5-plus & 70.00 & 4.80 & 32.15 \\
\bottomrule
\end{tabular*}
\caption{Active perception}
\label{tab:sub_b}
\end{subtable}

% ================= (c) 占满 100% 宽度 =================
\begin{subtable}[t]{\linewidth}
\centering
\begin{tabular*}{\linewidth}{@{\extracolsep{\fill}}lcccc@{}}
\toprule
\textbf{Models} & \textbf{Appropriate} & \textbf{Inappropriate} & \textbf{Fail} & \textbf{Time} \\
\midrule
gpt5.1 & 30.35 & 25.00 & 44.64 & 118.38 \\
gemini-3-flash & 57.89 & 26.31 & \textbf{15.78} & 33.76 \\
gemini-3.1-pro & \textbf{64.91} & \textbf{14.03} & 21.05 & 35.12 \\
qwen3.5-plus & 17.54 & 15.78 & 66.67 & \textbf{33.03} \\
\bottomrule
\end{tabular*}
\caption{Waypoints generation}
\label{tab:sub_c}
\end{subtable}
\end{table}

\subsection{How is the spatial intelligence of VLMs?}
Table~\ref{tab:benchmark} highlights a distinct spatial intelligence gap in current VLMs. While they excel at coarse, relational spatial reasoning (Tasks 1 and 2), they struggle with fine-grained metric perception from 2D observations (Task 3). This validates our framework: by decoupling VLM-based high-level planning from fine-grained action generation, we leverage their perceptual strengths while ensuring precise, generalizable locomotion and manipulation.

\section{Conclusion}
In this work, we study spatial-aware humanoid whole-body manipulation in complex 3D environments, which requires strong spatial understanding and robust action generalization. To address these challenges, we propose a generalizable humanoid loco-manipulation framework composed of two components: an Active Spatial Brain for active spatial perception, task planning, and subtask decomposition, and a Generalizable Action Cerebellum for executable action generation without task-specific real-robot data. Experiments on diverse spatial manipulation tasks demonstrate strong performance in both spatial perception and real-robot task execution across varied environments. Our results highlight the potential of integrating large-model-driven active spatial perception and decision making for generalizable humanoid manipulation.

% \clearpage\mbox{}Page \thepage\ of the manuscript.
% \clearpage\mbox{}Page \thepage\ of the manuscript. This is the last page.
% Now we have reached the maximum length of an ECCV \ECCVyear{} submission (excluding references and acknowledgements).
% References should start immediately after the main text, but can continue past p.\ 14 if needed. 
% \clearpage  % TODO FINAL: This \clearpage needs to be removed from both review and camera-ready versions.

\section*{Acknowledgements}
This work was supported partially by NSFC(92470202), Guangdong NSF Project (No. 2023B1515040025), Guangdong Key Research and Development Program (No.2024B0101040004, No. 2025B0909020002).

\bibliographystyle{splncs04}
\bibliography{main}

@inproceedings{homerobot,
  title={HomeRobot: Open-Vocabulary Mobile Manipulation},
  author={Yenamandra, Sriram and Ramachandran, Arun and Yadav, Karmesh and Wang, Austin S and Khanna, Mukul and Gervet, Theophile and Yang, Tsung-Yen and Jain, Vidhi and Clegg, Alexander and Turner, John M and others},
  booktitle={Annual Conference on Robot Learning},
  year={2023}
}

@article{aloha,
  title={Mobile aloha: Learning bimanual mobile manipulation with low-cost whole-body teleoperation},
  author={Fu, Zipeng and Zhao, Tony Z and Finn, Chelsea},
  journal={arXiv preprint arXiv:2401.02117},
  year={2024}
}

@inproceedings{momanipvla,
  title={Momanipvla: Transferring vision-language-action models for general mobile manipulation},
  author={Wu, Zhenyu and Zhou, Yuheng and Xu, Xiuwei and Wang, Ziwei and Yan, Haibin},
  booktitle={Proceedings of the Computer Vision and Pattern Recognition Conference},
  year={2025}
}

@article{Quad3,
  title={Learning whole-body manipulation for quadrupedal robot},
  author={Jeon, Seunghun and Jung, Moonkyu and Choi, Suyoung and Kim, Beomjoon and Hwangbo, Jemin},
  journal={IEEE Robotics and Automation Letters},
  year={2023},
}

@article{human2locoman,
  title={Human2locoman: Learning versatile quadrupedal manipulation with human pretraining},
  author={Niu, Yaru and Zhang, Yunzhe and Yu, Mingyang and Lin, Changyi and Li, Chenhao and Wang, Yikai and Yang, Yuxiang and Yu, Wenhao and Zhang, Tingnan and Li, Zhenzhen and others},
  journal={arXiv preprint arXiv:2506.16475},
  year={2025}
}

@inproceedings{QuadWBG,
  title={Quadwbg: Generalizable quadrupedal whole-body grasping},
  author={Wang, Jilong and Rajabov, Javokhirbek and Xu, Chaoyi and Zheng, Yiming and Wang, He},
  booktitle={IEEE International Conference on Robotics and Automation},
  year={2025}
}

@article{GR00t,
  title={Gr00t n1: An open foundation model for generalist humanoid robots},
  author={Bjorck, Johan and Casta{\~n}eda, Fernando and Cherniadev, Nikita and Da, Xingye and Ding, Runyu and Fan, Linxi and Fang, Yu and Fox, Dieter and Hu, Fengyuan and Huang, Spencer and others},
  journal={arXiv preprint arXiv:2503.14734},
  year={2025}
}

@article{egovla,
  title={Egovla: Learning vision-language-action models from egocentric human videos},
  author={Yang, Ruihan and Yu, Qinxi and Wu, Yecheng and Yan, Rui and Li, Borui and Cheng, An-Chieh and Zou, Xueyan and Fang, Yunhao and Cheng, Xuxin and Qiu, Ri-Zhao and others},
  journal={arXiv preprint arXiv:2507.12440},
  year={2025}
}

@inproceedings{humanoidpolicy,
  title={Humanoid Policy\~{} Human Policy},
  author={Qiu, Ri-Zhao and Yang, Shiqi and Cheng, Xuxin and Chawla, Chaitanya and Li, Jialong and He, Tairan and Yan, Ge and Yoon, David J and Hoque, Ryan and Paulsen, Lars and others},
  booktitle={Conference on Robot Learning},
  year={2025}
}

@article{agibot,
  title={Agibot world colosseo: A large-scale manipulation platform for scalable and intelligent embodied systems},
  author={Bu, Qingwen and Cai, Jisong and Chen, Li and Cui, Xiuqi and Ding, Yan and Feng, Siyuan and Gao, Shenyuan and He, Xindong and Hu, Xuan and Huang, Xu and others},
  journal={arXiv preprint arXiv:2503.06669},
  year={2025}
}

@inproceedings{okami,
  title={OKAMI: Teaching Humanoid Robots Manipulation Skills through Single Video Imitation},
  author={Li, Jinhan and Zhu, Yifeng and Xie, Yuqi and Jiang, Zhenyu and Seo, Mingyo and Pavlakos, Georgios and Zhu, Yuke},
  booktitle={Conference on Robot Learning},
  year={2024}
}

@article{omniretarget,
  title={Omniretarget: Interaction-preserving data generation for humanoid whole-body loco-manipulation and scene interaction},
  author={Yang, Lujie and Huang, Xiaoyu and Wu, Zhen and Kanazawa, Angjoo and Abbeel, Pieter and Sferrazza, Carmelo and Liu, C Karen and Duan, Rocky and Shi, Guanya},
  journal={arXiv preprint arXiv:2509.26633},
  year={2025}
}

@inproceedings{omnih2o,
  title={OmniH2O: Universal and Dexterous Human-to-Humanoid Whole-Body Teleoperation and Learning},
  author={He, Tairan and Luo, Zhengyi and He, Xialin and Xiao, Wenli and Zhang, Chong and Zhang, Weinan and Kitani, Kris M and Liu, Changliu and Shi, Guanya},
  booktitle={Conference on Robot Learning},
  year={2024}
}

@article{HumanoidVLA,
  title={Humanoid-vla: Towards universal humanoid control with visual integration},
  author={Ding, Pengxiang and Ma, Jianfei and Tong, Xinyang and Zou, Binghong and Luo, Xinxin and Fan, Yiguo and Wang, Ting and Lu, Hongchao and Mo, Panzhong and Liu, Jinxin and others},
  journal={arXiv preprint arXiv:2502.14795},
  year={2025}
}

@article{HOMIE,
  title={Homie: Humanoid loco-manipulation with isomorphic exoskeleton cockpit},
  author={Ben, Qingwei and Jia, Feiyu and Zeng, Jia and Dong, Junting and Lin, Dahua and Pang, Jiangmiao},
  journal={arXiv preprint arXiv:2502.13013},
  year={2025}
}

@inproceedings{Twist,
  title={TWIST: Teleoperated Whole-Body Imitation System},
  author={Ze, Yanjie and Chen, Zixuan and Araujo, Joao Pedro and Cao, Zi-ang and Peng, Xue Bin and Wu, Jiajun and Liu, Karen},
  booktitle={Conference on Robot Learning},
  year= {2025},
}

@article{HMI,
  title={Humanoid Manipulation Interface: Humanoid Whole-Body Manipulation from Robot-Free Demonstrations},
  author={Nai, Ruiqian and Zheng, Boyuan and Zhao, Junming and Zhu, Haodong and Dai, Sicong and Chen, Zunhao and Hu, Yihang and Hu, Yingdong and Zhang, Tong and Wen, Chuan and others},
  journal={arXiv preprint arXiv:2602.06643},
  year={2026}
}

@inproceedings{H2O,
  title={Learning human-to-humanoid real-time whole-body teleoperation},
  author={He, Tairan and Luo, Zhengyi and Xiao, Wenli and Zhang, Chong and Kitani, Kris and Liu, Changliu and Shi, Guanya},
  booktitle={IEEE International Conference on Intelligent Robots and Systems},
  year={2024}
}

@inproceedings{CLONE,
  title={Clone: Closed-loop whole-body humanoid teleoperation for long-horizon tasks},
  author={Li, Yixuan and Lin, Yutang and Cui, Jieming and Liu, Tengyu and Liang, Wei and Zhu, Yixin and Huang, Siyuan},
  booktitle={Conference on Robot Learning},
  year={2025}
}

@article{WBVLA,
  title={Wholebodyvla: Towards unified latent vla for whole-body loco-manipulation control},
  author={Jiang, Haoran and Chen, Jin and Bu, Qingwen and Chen, Li and Shi, Modi and Zhang, Yanjie and Li, Delong and Suo, Chuanzhe and Wang, Chuang and Peng, Zhihui and others},
  journal={arXiv preprint arXiv:2512.11047},
  year={2025}
}

@inproceedings{iDP3,
  title={Generalizable humanoid manipulation with 3d diffusion policies},
  author={Ze, Yanjie and Chen, Zixuan and Wang, Wenhao and Chen, Tianyi and He, Xialin and Yuan, Ying and Peng, Xue Bin and Wu, Jiajun},
  booktitle={IEEE International Conference on Intelligent Robots and Systems},
  year={2025}
}

@article{Being0,
  title={Being-0: A humanoid robotic agent with vision-language models and modular skills},
  author={Yuan, Haoqi and Bai, Yu and Fu, Yuhui and Zhou, Bohan and Feng, Yicheng and Xu, Xinrun and Zhan, Yi and Karlsson, B{\"o}rje F and Lu, Zongqing},
  journal={arXiv preprint arXiv:2503.12533},
  year={2025}
}

@article{LeVERB,
  title={Leverb: Humanoid whole-body control with latent vision-language instruction},
  author={Xue, Haoru and Huang, Xiaoyu and Niu, Dantong and Liao, Qiayuan and Kragerud, Thomas and Gravdahl, Jan Tommy and Peng, Xue Bin and Shi, Guanya and Darrell, Trevor and Sreenath, Koushil and others},
  journal={arXiv preprint arXiv:2506.13751},
  year={2025}
}

@article{Trajbooster,
  title={TrajBooster: Boosting Humanoid Whole-Body Manipulation via Trajectory-Centric Learning},
  author={Liu, Jiacheng and Ding, Pengxiang and Zhou, Qihang and Wu, Yuxuan and Huang, Da and Peng, Zimian and Xiao, Wei and Zhang, Weinan and Yang, Lixin and Lu, Cewu and others},
  journal={arXiv preprint arXiv:2509.11839},
  year={2025}
}

@article{ResMimic,
  title={Resmimic: From general motion tracking to humanoid whole-body loco-manipulation via residual learning},
  author={Zhao, Siheng and Ze, Yanjie and Wang, Yue and Liu, C Karen and Abbeel, Pieter and Shi, Guanya and Duan, Rocky},
  journal={arXiv preprint arXiv:2510.05070},
  year={2025}
}

@article{R2S2,
  title={Unleashing humanoid reaching potential via real-world-ready skill space},
  author={Zhang, Zhikai and Chen, Chao and Xue, Han and Wang, Jilong and Liang, Sikai and Liu, Yun and Zhang, Zongzhang and Wang, He and Yi, Li},
  journal={IEEE Robotics and Automation Letters},
  year={2025}
}

@article{ULC,
  title={Ulc: A unified and fine-grained controller for humanoid loco-manipulation},
  author={Sun, Wandong and Feng, Luying and Cao, Baoshi and Liu, Yang and Jin, Yaochu and Xie, Zongwu},
  journal={arXiv preprint arXiv:2507.06905},
  year={2025}
}

@article{FALCON,
  title={Falcon: Learning force-adaptive humanoid loco-manipulation},
  author={Zhang, Yuanhang and Yuan, Yifu and Gurunath, Prajwal and Gupta, Ishita and Omidshafiei, Shayegan and Agha-mohammadi, Ali-akbar and Vazquez-Chanlatte, Marcell and Pedersen, Liam and He, Tairan and Shi, Guanya},
  journal={arXiv preprint arXiv:2505.06776},
  year={2025}
}

@article{AMO,
  title={Amo: Adaptive motion optimization for hyper-dexterous humanoid whole-body control},
  author={Li, Jialong and Cheng, Xuxin and Huang, Tianshu and Yang, Shiqi and Qiu, Ri-Zhao and Wang, Xiaolong},
  journal={arXiv preprint arXiv:2505.03738},
  year={2025}
}

@article{clot,
  title={CLOT: Closed-Loop Global Motion Tracking for Whole-Body Humanoid Teleoperation},
  author={Zhu, Tengjie and Cai, Guanyu and Zhaohui, Yang and Ren, Guanzhu and Xie, Haohui and Wang, ZiRui and Wu, Junsong and Wang, Jingbo and Yang, Xiaokang and Mu, Yao and others},
  journal={arXiv preprint arXiv:2602.15060},
  year={2026}
}

@article{VIRAL,
  title={VIRAL: Visual Sim-to-Real at Scale for Humanoid Loco-Manipulation},
  author={He, Tairan and Wang, Zi and Xue, Haoru and Ben, Qingwei and Luo, Zhengyi and Xiao, Wenli and Yuan, Ye and Da, Xingye and Casta{\~n}eda, Fernando and Sastry, Shankar and others},
  journal={arXiv preprint arXiv:2511.15200},
  year={2025}
}

@inproceedings{DoasIcan,
  author       = {Brian Ichter and
                  Anthony Brohan and
                  Yevgen Chebotar and
                  Chelsea Finn and
                  Karol Hausman and
                  Alexander Herzog and
                  Daniel Ho and
                  Julian Ibarz and
                  Alex Irpan and
                  Eric Jang and
                  Ryan Julian and
                  Dmitry Kalashnikov and
                  Sergey Levine and
                  Yao Lu and
                  Carolina Parada and
                  Kanishka Rao and
                  Pierre Sermanet and
                  Alexander Toshev and
                  Vincent Vanhoucke and
                  Fei Xia and
                  Ted Xiao and
                  Peng Xu and
                  Mengyuan Yan and
                  Noah Brown and
                  Michael Ahn and
                  Omar Cortes and
                  Nicolas Sievers and
                  Clayton Tan and
                  Sichun Xu and
                  Diego Reyes and
                  Jarek Rettinghouse and
                  Jornell Quiambao and
                  Peter Pastor and
                  Linda Luu and
                  Kuang{-}Huei Lee and
                  Yuheng Kuang and
                  Sally Jesmonth and
                  Nikhil J. Joshi and
                  Kyle Jeffrey and
                  Rosario Jauregui Ruano and
                  Jasmine Hsu and
                  Keerthana Gopalakrishnan and
                  Byron David and
                  Andy Zeng and
                  Chuyuan Kelly Fu},
  title        = {Do As {I} Can, Not As {I} Say: Grounding Language in Robotic Affordances},
  booktitle    = {Conference on Robot Learning},
  year         = {2022}
}

@inproceedings{InnerMono,
  author       = {Wenlong Huang and
                  Fei Xia and
                  Ted Xiao and
                  Harris Chan and
                  Jacky Liang and
                  Pete Florence and
                  Andy Zeng and
                  Jonathan Tompson and
                  Igor Mordatch and
                  Yevgen Chebotar and
                  Pierre Sermanet and
                  Tomas Jackson and
                  Noah Brown and
                  Linda Luu and
                  Sergey Levine and
                  Karol Hausman and
                  Brian Ichter},
  title        = {Inner Monologue: Embodied Reasoning through Planning with Language
                  Models},
  booktitle    = {Conference on Robot Learning},
  year         = {2022}
}

@inproceedings{ProgPrompt,
  author       = {Ishika Singh and
                  Valts Blukis and
                  Arsalan Mousavian and
                  Ankit Goyal and
                  Danfei Xu and
                  Jonathan Tremblay and
                  Dieter Fox and
                  Jesse Thomason and
                  Animesh Garg},
  title        = {ProgPrompt: Generating Situated Robot Task Plans using Large Language Models},
  booktitle    = {IEEE International Conference on Robotics and Automation},
  year         = {2023}
}

@inproceedings{Voxposer,
  author       = {Wenlong Huang and
                  Chen Wang and
                  Ruohan Zhang and
                  Yunzhu Li and
                  Jiajun Wu and
                  Li Fei{-}Fei},
  title        = {VoxPoser: Composable 3D Value Maps for Robotic Manipulation with Language Models},
  booktitle    = {Conference on Robot Learning},
  year         = {2023}
}

@inproceedings{Rekep,
  author       = {Wenlong Huang and
                  Chen Wang and
                  Yunzhu Li and
                  Ruohan Zhang and
                  Li Fei{-}Fei},
  title        = {ReKep: Spatio-Temporal Reasoning of Relational Keypoint Constraints
                  for Robotic Manipulation},
  booktitle    = {Conference on Robot Learning},
  year         = {2024}
}

@article{ReSem3D,
  author={Su, Chenyu and Shang, Weiwei and Qian, Chen and Zhang, Fei and Cong, Shuang},
  title        = {ReSemAct: Advancing Fine-Grained Robotic Manipulation via Semantic Structuring and Affordance Refinement},
  journal      ={arXiv preprint arXiv:2507.18262},
  year         = {2025}
}

@inproceedings{omnimanip,
  author={Pan, Mingjie and Zhang, Jiyao and Wu, Tianshu and Zhao, Yinghao and Gao, Wenlong and Dong, Hao},
  title={Omnimanip: Towards general robotic manipulation via object-centric interaction primitives as spatial constraints},
  booktitle={Proceedings of the Computer Vision and Pattern Recognition Conference},
  year={2025}
}

@inproceedings{LLMfewshot,
  author       = {Tom B. Brown and
                  Benjamin Mann and
                  Nick Ryder and
                  Melanie Subbiah and
                  Jared Kaplan and
                  Prafulla Dhariwal and
                  Arvind Neelakantan and
                  Pranav Shyam and
                  Girish Sastry and
                  Amanda Askell and
                  Sandhini Agarwal and
                  Ariel Herbert{-}Voss and
                  Gretchen Krueger and
                  Tom Henighan and
                  Rewon Child and
                  Aditya Ramesh and
                  Daniel M. Ziegler and
                  Jeffrey Wu and
                  Clemens Winter and
                  Christopher Hesse and
                  Mark Chen and
                  Eric Sigler and
                  Mateusz Litwin and
                  Scott Gray and
                  Benjamin Chess and
                  Jack Clark and
                  Christopher Berner and
                  Sam McCandlish and
                  Alec Radford and
                  Ilya Sutskever and
                  Dario Amodei},
  title        = {Language Models are Few-Shot Learners},
  booktitle    = {Advances in Neural Information Processing Systems},
  year         = {2020}
}

@inproceedings{LLMzeroshot,
  author       = {Takeshi Kojima and
                  Shixiang Shane Gu and
                  Machel Reid and
                  Yutaka Matsuo and
                  Yusuke Iwasawa},
  title        = {Large Language Models are Zero-Shot Reasoners},
  booktitle    = {Advances in Neural Information Processing Systems},
  year         = {2022}
}

@inproceedings{LLMCode,
  author       = {Aman Madaan and
                  Shuyan Zhou and
                  Uri Alon and
                  Yiming Yang and
                  Graham Neubig},
  title        = {Language Models of Code are Few-Shot Commonsense Learners},
  booktitle    = {Proceedings of the Conference on Empirical Methods in Natural
                  Language Processing},
  year         = {2022}
}

@article{gptcode,
  author={Dong, Yihong and Jiang, Xue and Qian, Jiaru and Wang, Tian and Zhang, Kechi and Jin, Zhi and Li, Ge},
  title={A survey on code generation with llm-based agents},
  journal={arXiv preprint arXiv:2508.00083},
  year={2025}
}

@article{SpatialActor,
    title={SpatialActor: Exploring Disentangled Spatial Representations for Robust Robotic Manipulation},
    author={Shi, Hao and Xie, Bin and Liu, Yingfei and Yue, Yang and Wang, Tiancai and Fan, Haoqiang and Zhang, Xiangyu and Huang, Gao},
    journal={arXiv preprint arXiv:2511.09555},
    year={2025}
}

@INPROCEEDINGS{CoPa,
  author={Huang, Haoxu and Lin, Fanqi and Hu, Yingdong and Wang, Shengjie and Gao, Yang},
  booktitle={IEEE International Conference on Intelligent Robots and Systems}, 
  title={CoPa: General Robotic Manipulation through Spatial Constraints of Parts with Foundation Models}, 
  year={2024}
}

@inproceedings{SOFAR,
  title={SoFar: Language-Grounded Orientation Bridges Spatial Reasoning and Object Manipulation},
  author={Qi, Zekun and Zhang, Wenyao and Ding, Yufei and Dong, Runpei and Yu, XinQiang and Li, Jingwen and Xu, Lingyun and Li, Baoyu and He, Xialin and Fan, Guofan and others},
  booktitle={Advances in Neural Information Processing Systems},
  year={2025}
}

@article{internvlam1,
  title   = {InternVLA-M1: A Spatially Guided Vision-Language-Action Framework for Generalist Robot Policy},
  author  = {InternVLA-M1 Contributors},
  journal = {arXiv preprint arXiv:2510.13778},
  year    = {2025}
}

@article{ActiveVLA,
  title={ActiveVLA: Injecting Active Perception into Vision-Language-Action Models for Precise 3D Robotic Manipulation},
  author={Liu, Zhenyang and Gu, Yongchong and Wang, Yikai and Xue, Xiangyang and Fu, Yanwei},
  journal={arXiv preprint arXiv:2601.08325},
  year={2026}
}

@article{SpatialPolicy,
  title={Spatial policy: Guiding visuomotor robotic manipulation with spatial-aware modeling and reasoning},
  author={Liu, Yijun and Liu, Yuwei and Meng, Yuan and Zhang, Jieheng and Zhou, Yuwei and Li, Ye and Jiang, Jiacheng and Ji, Kangye and Ge, Shijia and Wang, Zhi and others},
  journal={arXiv preprint arXiv:2508.15874},
  year={2025}
}

@article{RoboPoint,
  title={RoboPoint: A Vision-Language Model for Spatial Affordance Prediction for Robotics},
  author={Yuan, Wentao and Duan, Jiafei and Blukis, Valts and Pumacay, Wilbert and Krishna, Ranjay and Murali, Adithyavairavan and Mousavian, Arsalan and Fox, Dieter},
  journal={arXiv preprint arXiv:2406.10721},
  year={2024}
}

@article{MOKA,
      title={MOKA: Open-World Robotic Manipulation through Mark-Based Visual Prompting},
      author={Kuan Fang and Fangchen Liu and Pieter Abbeel and Sergey Levine},
      journal={Robotics: Science and Systems},
      year={2024}
  }

@article{AimBot,
  title={AimBot: A Simple Auxiliary Visual Cue to Enhance Spatial Awareness of Visuomotor Policies},
  author={Yinpei Dai and Jayjun Lee and Yichi Zhang and Ziqiao Ma and Jed Yang and Amir Zadeh and Chuan Li and Nima Fazeli and Joyce Chai},
  journal={Conference on Robot Learning},
  year={2025},
}

@inproceedings{todorov2012mujoco,
  title={Mujoco: A physics engine for model-based control},
  author={Todorov, Emanuel and Erez, Tom and Tassa, Yuval},
  booktitle={2012 IEEE/RSJ international conference on intelligent robots and systems},
  pages={5026--5033},
  year={2012},
  organization={IEEE}
}

@misc{hyper3d2024,
  author       = {Hyper3D},
  title        = {Hyper3D: AI-Powered 3D Model Generator},
  year         = {2024},
  url          = {https://hyper3d.ai/},
  note         = {Accessed: 2026-02-28},
}

@inproceedings{liu2025easyhoi,
  title={Easyhoi: Unleashing the power of large models for reconstructing hand-object interactions in the wild},
  author={Liu, Yumeng and Long, Xiaoxiao and Yang, Zemin and Liu, Yuan and Habermann, Marc and Theobalt, Christian and Ma, Yuexin and Wang, Wenping},
  booktitle={Proceedings of the Computer Vision and Pattern Recognition Conference},
  pages={7037--7047},
  year={2025}
}

@inproceedings{psi0,
  author       = {Songlin Wei and
                  Hongyi Jing and
                  Boqian Li and
                  Zhenyu Zhao and
                  Jiageng Mao and
                  Zhenhao Ni and
                  Sicheng He and
                  Jie Liu and
                  Xiawei Liu and
                  Kaidi Kang and
                  Sheng Zang and
                  Weiduo Yuan and
                  Marco Pavone and
                  Di Huang and
                  Yue Wang},
  title        = {$\Psi_0$: An Open Foundation Model Towards Universal Humanoid Loco-Manipulation},
  booktitle    = {Robotics: Science and Systems},
  year         = {2026}
}

@inproceedings{CaP,
  author       = {Jacky Liang and
                  Wenlong Huang and
                  Fei Xia and
                  Peng Xu and
                  Karol Hausman and
                  Brian Ichter and
                  Pete Florence and
                  Andy Zeng},
  title        = {Code as Policies: Language Model Programs for Embodied Control},
  booktitle    = {IEEE International Conference on Robotics and Automation},
  year         = {2023}
}

@inproceedings{wei2025afforddexgrasp,
  title={Afforddexgrasp: Open-set language-guided dexterous grasp with generalizable-instructive affordance},
  author={Wei, Yi-Lin and Lin, Mu and Lin, Yuhao and Jiang, Jian-Jian and Wu, Xiao-Ming and Zeng, Ling-An and Zheng, Wei-Shi},
  booktitle={Proceedings of the IEEE/CVF International Conference on Computer Vision},
  pages={11818--11828},
  year={2025}
}

@article{lin2026bidexgrasp,
  title={BiDexGrasp: Coordinated Bimanual Dexterous Grasps across Object Geometries and Sizes},
  author={Lin, Mu and Wei, Yi-Lin and Chen, Jiaxuan and Lin, Yuhao and Chen, Shuoyu and Lyu, Jiangran and Chen, Jiayi and Tang, Yansong and Wang, He and Zheng, Wei-Shi},
  journal={arXiv preprint arXiv:2604.06589},
  year={2026}
}

@inproceedings{zhou2025mitigating,
  title={Mitigating the human-robot domain discrepancy in visual pre-training for robotic manipulation},
  author={Zhou, Jiaming and Ma, Teli and Lin, Kun-Yu and Wang, Zifan and Qiu, Ronghe and Liang, Junwei},
  booktitle={Proceedings of the computer vision and pattern recognition conference},
  pages={22551--22561},
  year={2025}
}

@inproceedings{huang2026beyond,
  title={Beyond Mimicry: Learning Whole-Body Human-Humanoid Interaction from Human-Human Demonstrations},
  author={Huang, Wei-Jin and Zhang, Yue-Yi and Wei, Yi-Lin and Xia, Zhi-Wei and Tan, Juantao and Li, Yuan-Ming and Zhao, Zhilin and Zheng, Wei-Shi},
  booktitle={Proceedings of the IEEE/CVF Conference on Computer Vision and Pattern Recognition},
  pages={30740--30749},
  year={2026}
}

@article{zhou2026exploring,
  title={Exploring the limits of vision-language-action manipulation in cross-task generalization},
  author={Zhou, Jiaming and Ye, Ke and Liu, Jiayi and Ma, Teli and Wang, Zifan and Qiu, Ronghe and Lin, Kun-Yu and Zhao, Zhilin and Liang, Junwei},
  journal={Advances in Neural Information Processing Systems},
  volume={38},
  pages={139899--139927},
  year={2026}
}

@article{jiang2026task,
  title={Task Editing for Generalizable 3D Visuomotor Policy Learning},
  author={Jiang, Jian-Jian and Yang, YiHan and Wei, Lan and Luo, Yuming and Wu, Xiao-Ming and Chen, Xuhang and Fan, Bin and Zhang, Dandan and Zheng, Wei-Shi},
  journal={arXiv preprint arXiv:2606.07012},
  year={2026}
}

@article{xue2025demogen,
  title={Demogen: Synthetic demonstration generation for data-efficient visuomotor policy learning},
  author={Xue, Zhengrong and Deng, Shuying and Chen, Zhenyang and Wang, Yixuan and Yuan, Zhecheng and Xu, Huazhe},
  journal={arXiv preprint arXiv:2502.16932},
  year={2025}
}

@inproceedings{wei2026omnidexgrasp,
  title={OmniDexGrasp: Generalizable Dexterous Grasping via Foundation Model and Force Feedback},
  author={Wei, Yi-Lin and Luo, Zhexi and Lin, Yuhao and Lin, Mu and Liang, Zhizhao and Chen, Shuoyu and Zheng, Wei-Shi},
  booktitle={IEEE International Conference on Robotics and Automation (ICRA)},
  year={2026}
}

@inproceedings{fang2020graspnet,
  title={Graspnet-1billion: A large-scale benchmark for general object grasping},
  author={Fang, Hao-Shu and Wang, Chenxi and Gou, Minghao and Lu, Cewu},
  booktitle={Proceedings of the IEEE/CVF conference on computer vision and pattern recognition},
  pages={11444--11453},
  year={2020}
}

@inproceedings{jiang2025rethinking,
  title={Rethinking bimanual robotic manipulation: Learning with decoupled interaction framework},
  author={Jiang, Jian-Jian and Wu, Xiao-Ming and He, Yi-Xiang and Zeng, Ling-An and Wei, Yi-Lin and Zhang, Dandan and Zheng, Wei-Shi},
  booktitle={Proceedings of the IEEE/CVF International Conference on Computer Vision},
  pages={12427--12437},
  year={2025}
}

@inproceedings{wang2025task,
  title={Task-oriented 6-dof grasp pose detection in clutters},
  author={Wang, An-Lan and Chen, Nuo and Lin, Kun-Yu and Li, Yuan-Ming and Zheng, Wei-Shi},
  booktitle={2025 IEEE International Conference on Robotics and Automation (ICRA)},
  pages={5692--5698},
  year={2025},
  organization={IEEE}
}

@inproceedings{wu2024economic,
  title={An economic framework for 6-dof grasp detection},
  author={Wu, Xiao-Ming and Cai, Jia-Feng and Jiang, Jian-Jian and Zheng, Dian and Wei, Yi-Lin and Zheng, Wei-Shi},
  booktitle={European Conference on Computer Vision},
  pages={357--375},
  year={2024},
  organization={Springer}
}

@inproceedings{xu2024dexterous,
  title={Dexterous grasp transformer},
  author={Xu, Guo-Hao and Wei, Yi-Lin and Zheng, Dian and Wu, Xiao-Ming and Zheng, Wei-Shi},
  booktitle={Proceedings of the IEEE/CVF Conference on Computer Vision and Pattern Recognition},
  pages={17933--17942},
  year={2024}
}

@article{wei2024grasp,
  title={Grasp as you say: Language-guided dexterous grasp generation},
  author={Wei, Yi-Lin and Jiang, Jian-Jian and Xing, Chengyi and Tan, Xian-Tuo and Wu, Xiao-Ming and Li, Hao and Cutkosky, Mark and Zheng, Wei-Shi},
  journal={Advances in Neural Information Processing Systems},
  volume={37},
  pages={46881--46907},
  year={2024}
}
\end{document}